\documentclass[sigconf]{aamas}

\usepackage{algorithm}
\usepackage{algorithmic}
\usepackage{tikz}
\usetikzlibrary{positioning,arrows.meta}
\definecolor{gwrong}{RGB}{150,0,0}
\definecolor{gok}{RGB}{0,100,0}
\definecolor{ourmethodblue}{RGB}{222,235,247}
\newcommand{\ourrowcolor}{\noalign{\begingroup\color{ourmethodblue}\hrule height 2.55ex\endgroup\kern-2.55ex}}
\makeatletter
\gdef\@copyrightpermission{
  \begin{minipage}{0.2\columnwidth}
   \href{https://creativecommons.org/licenses/by/4.0/}{\includegraphics[width=0.90\textwidth]{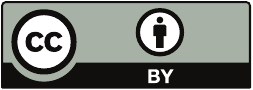}}
  \end{minipage}\hfill
  \begin{minipage}{0.8\columnwidth}
    \href{https://creativecommons.org/licenses/by/4.0/}{This work is licensed under a Creative Commons Attribution International 4.0 License.}
  \end{minipage}
  \vspace{5pt}
}
\makeatother

\setcopyright{ifaamas}
\acmConference[AAMAS 2027]{Proc.\@ of the 26th International Conference
on Autonomous Agents and Multiagent Systems (AAMAS 2027)}{3--7 May 2027}{Hanoi, Vietnam}{M.~Baldoni, F.~Fang, W.~Yeoh, N.~Yorke-Smith (eds.)}
\copyrightyear{2027}
\acmYear{2027}
\acmDOI{}
\acmISBN{}

\submissionType{Research Paper Track}

\title[Value-Conditioned Proposal Acquisition]{Value-Conditioned Proposal Acquisition for Test-Time Reasoning}

\author{Zeyan Li}
\affiliation{%
  \institution{Shanghai Jiao Tong University}
  \city{Shanghai}
  \country{China}}

\author{Jianfeng Xu}
\affiliation{%
  \institution{Shanghai Jiao Tong University}
  \city{Shanghai}
  \country{China}}

\begin{abstract}
Sampling and selection spend test-time compute inside a candidate pool, but they cannot recover a correct answer that never enters it. We introduce ValueFormation, a test-time policy that decides after each primary-model draw whether to stop, continue sampling, or acquire one proposal from a heterogeneous reasoning specialist. The three actions are compared on one cost-regularized value scale learned from vote-state features; the calibration is frozen across tasks and primary models, and acquired proposals are accepted only through a frozen posterior comparison. On held-out MATH-500, one Llama-fitted calibration improves over the strongest of eight identically evaluated baselines by more than fifteen points for each of three 8B primaries and transfers unchanged to Qwen3-14B. Removing acquisition lowers Llama MATH accuracy by 20.43 points, while same-trace and same-access replays attribute the gain to state-conditioned acquisition rather than specialist access alone. Where primary coverage is high, as on GSM8K, acquisition is rare and gains mainly come from cheaper allocation; on MBPP, a fixed code route attains the top accuracy at the lowest interaction cost. Varying the token price monotonically reduces calls without retraining.
\end{abstract}



\begin{document}

\maketitle


\section{Introduction}

Language agents solve harder problems when they are allowed to spend more computation at inference time \cite{snell2024testtime}. Most of that budget is spent inside one model: it samples several solutions and selects among them by voting \cite{wang2023selfconsistency} or with a trained verifier \cite{cobbe2021math}. A multi-agent alternative is to recruit a differently trained collaborator that contributes its own proposal \cite{du2024debate,li2024moreagents}. The two currencies are rarely priced against each other: sampling research makes each additional call from the same model cheaper or better timed, while collaboration research decides whom to invoke without watching how the current attempt is unfolding. This paper treats them as one allocation problem for a single agent under a shared token budget.

\begin{figure}[t]
\centering
\includegraphics[width=\columnwidth]{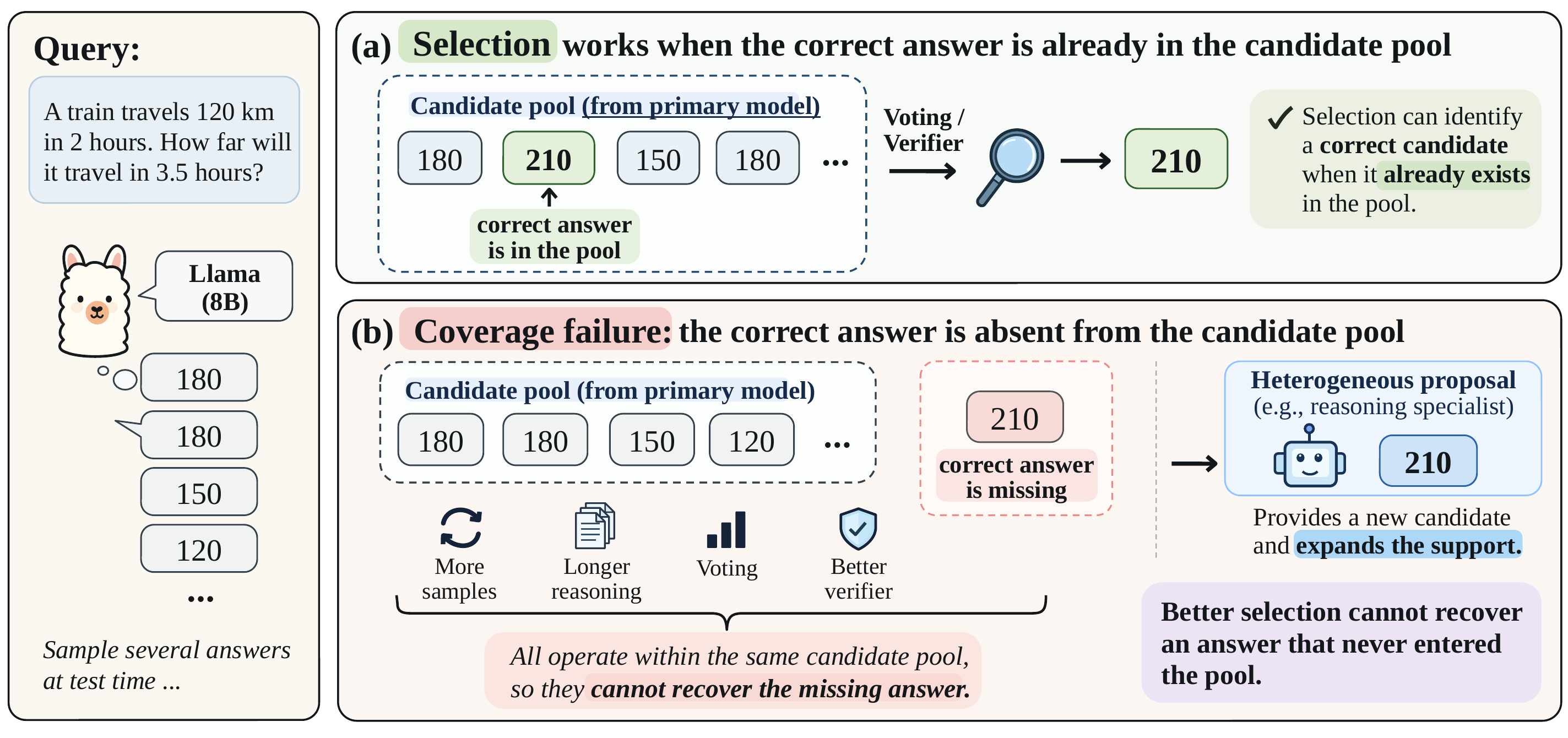}
\caption{Selection versus coverage failure. Selection works when the correct answer is already in the primary candidate pool; a heterogeneous proposal is useful when the pool lacks that answer.}
\Description{Illustration of a math query where primary samples form a candidate pool. The top panel shows voting selecting the correct answer when it is present. The bottom panel shows a missing correct answer that same-pool operations cannot recover, while a heterogeneous specialist adds the missing candidate.}
\label{fig:coverage}
\end{figure}

Selection, however sophisticated, can only reweight what the current pool contains. Continued stochastic sampling can add new support, but repeated-sampling studies show that coverage on hard problems often grows slowly with budget \cite{brown2024monkeys}. Figure~\ref{fig:coverage} illustrates the distinction: voting or verification can identify a correct candidate when it is already present, but cannot recover one that never entered the pool. We call the latter a failure of candidate coverage, distinct from a failure to select a present answer. Coverage failures are difficult to diagnose from agreement alone: a confidently wrong pool can resemble a confidently correct one.

The two bottlenecks require different interventions. Better voting or verification addresses selection among present answers. A proposal from a differently specialized model can instead change the support of the candidate set. This distinction motivates a conditional policy: continue the primary when another draw is valuable, and acquire only when a heterogeneous proposal has higher expected utility.

Existing methods do not make this choice. Those that adapt \emph{how much} to spend act on the samples of a single model, whether by calibrated stopping \cite{huang2026cats}, budgeted decomposition \cite{lin2026planandbudget}, or vote--verifier aggregation \cite{zhou2026variation,kuang2026optimalaggregation,li2026brokencompass}. Those that choose \emph{which model} to call instead route from the question \cite{ong2025routellm} or escalate a cascade using confidence in an earlier response \cite{chen2023frugalgpt}; neither prices that call against another draw after observing an evolving vote pool. Multi-agent systems invite several models to collaborate \cite{du2024debate,li2024moreagents}, yet their apparent advantage can disappear when reasoning-token budgets are matched \cite{tran2026sas}.

Between these two lines sits the setting we study: a policy that watches one model's own samples, detects the state where more of the same will not help, and spends the remaining budget on a different kind of call. The open question is whether, under a fixed inference budget, the next call is better spent on one more correlated vote or on a proposal from a model with different capabilities.

Our method, ValueFormation, makes this decision on a common value scale. From the current vote histogram it predicts the success of stopping, continuing primary sampling to the cap, and acquiring a reasoning proposal. Each prediction is reduced by the action's expected token cost under one price $\lambda$, and the largest utility determines the next call. An acquired proposal is accepted only when frozen posterior evidence favors it over the incumbent leader. The predictors are fitted once on Llama development trajectories pooled across three mathematics tasks, receive no benchmark identity, and are transferred unchanged to Qwen and Ministral primaries. The public code interface selects a fixed code-specialist route for program-synthesis tasks.

The prospective results test both sides of the coverage account. On held-out MATH-500 rows, where primary coverage is scarce, ValueFormation improves over the strongest exact-ID baseline by 17.8 points with a Llama primary, 20.8 with Qwen, and 15.6 with Ministral, and removing acquisition from the Llama policy lowers accuracy from 80.3\% to 59.9\%. Same-trace and same-access replays attribute this gain to the state-conditioned choice rather than to specialist access: giving every alternative rule the identical payload, integrator, and continuation access still favors the policy by 12.3 points over matched-rate random acquisition, and its acquisitions concentrate on problems whose pool lacks the answer (85\% versus 41\%). An unconditional control that calls the same specialist once on every problem never exceeds the policy on MATH-500 and scores below it on GSM8K, where primary coverage is already high and the policy instead acquires rarely. The same frozen predictors transfer unchanged to a Qwen3-14B primary, and increasing $\lambda$ reduces realized calls and tokens on every task, as the utility formulation predicts.

Our contributions address these two gaps. First, we separate candidate coverage from candidate selection and formulate heterogeneous collaboration as an action in budgeted metareasoning, making the kind of the next call a decision variable rather than a fixed design choice. Second, we realize that formulation as one task-blind, cost-regularized action-value policy with a monotone cost comparative static; the policy decides from the evolving vote state after problem-specific evidence has accumulated, not in advance from the question alone. Third, we test the resulting predictions prospectively across four primary models: acquisition is load-bearing under a direct intervention, the state-conditioned rule beats every matched-access alternative where coverage is scarce, and the acceptance rule protects the policy under a deliberately weakened specialist.

\section{Related Work}

\paragraph{Test-time compute within one model.} The default way to spend inference compute is to sample several solutions from one model and select among them. Self-consistency votes over sampled chains \cite{wang2023selfconsistency}; adaptive stopping decides when the vote has settled and stops drawing \cite{aggarwal2023adaptive,li2024esc,wang2025dsc}; verifiers and process reward models rerank what was sampled \cite{cobbe2021math,lightman2024verify,zhang2025genrm}. The stopping statistics descend from sequential hypothesis testing \cite{wald1945}. The newest members of this family budget the same-model compute by calibrated confidence, difficulty-adaptive allocation, question decomposition, or principled vote--verifier aggregation \cite{huang2026cats,snell2024testtime,lin2026planandbudget,kuang2026optimalaggregation,zhou2026variation,li2026brokencompass}. These methods all allocate the same object: samples from one model distribution. Repeated-sampling measurements show that on hard problems the correct answer often never enters the pool at all \cite{brown2024monkeys}, and no reranker or stopping rule can select an answer that was never sampled. Our work keeps the budgeted-decision frame but changes what is being bought: not only further draws from the same distribution, but proposals from outside it.

\paragraph{Choosing among models and agents.} A separate literature chooses which model to call. Learned routers map each question directly to a model \cite{ong2025routellm,jitkrittum2026universal}, while cascades can escalate after scoring an earlier response \cite{chen2023frugalgpt}. Staged collaboration adapts execution through planners over executors \cite{lee2025cope}, adaptive cost-aware selection \cite{zhang2025lightrouter}, or roles invited to interact \cite{du2024debate,li2024moreagents,madaan2023selfrefine,shinn2023reflexion,zhuge2024redel,guo2024dylan}; runtime-efficiency work budgets the token spend of such systems explicitly \cite{lin2026stopwasting}, yet matched-budget studies show that multi-agent architectures need not outperform a single agent when reasoning tokens are held constant \cite{tran2026sas}. The classical pricing ingredients predate all of this: rational metareasoning prices a costly computation by its expected value \cite{russell1991}, and pooling correlated raters with fitted per-source reliabilities is Dawid--Skene \cite{dawid1979}. What remains uncombined is the local choice among stopping, another draw from the current model, and one heterogeneous proposal after a vote state has formed. Our policy prices exactly those three actions on the evidence accumulated for the current problem.

\paragraph{Program-aided and execution-grounded reasoning.} Generating and executing programs grounds answer selection in task-specific feedback: a candidate that runs and passes visible checks carries evidence that merely plausible text does not \cite{gao2023pal,chen2023pot,li2025sstar}, and voting over code candidates has its own selection dynamics \cite{launer2026majorityvoting}. Our code route uses this feedback where the public interface supplies it, through literal public examples and a fixed specialist route, and the mathematics experiments carry the central acquisition claim.

\section{Value-Conditioned Team Formation}\label{sec:vf-method}
ValueFormation treats collaboration as a metareasoning decision. After each primary-model draw, it chooses among three actions under one utility scale: stop with the current leader, continue primary sampling, or acquire one proposal from a reasoning-specialized model. The policy is calibrated once on disjoint development trajectories and receives no benchmark identity at inference time.

\subsection{Problem setting}\label{sec:problem}
\begin{definition}[Vote-state episode]\label{def:episode}
For a problem $x$ and primary model $M$, an episode draws responses from $M$ sequentially, up to a cap of $K=10$ draws and outer hard limits of $B_c$ calls and $B_t$ tokens. Each response is parsed and canonicalized to one answer identity or fails to parse; parsed answers accumulate in a vote histogram $\mathcal A$. After each draw $k\ge1$, the policy chooses on the basis of the vote histogram alone among returning the incumbent leader (\textsc{stop}), drawing once more from $M$ (\textsc{primary}), and terminating sampling to acquire one proposal from a reasoning specialist (\textsc{acquire}). The episode ends when the policy stops, acquires, or reaches the cap.
\end{definition}

\begin{definition}[Budgeted action policy]\label{def:policy}
Fix a token price $\lambda\ge0$. For a policy $\pi$ that returns answer $a_\pi(x)$ after spending $t_\pi(x)$ prompt-plus-completion tokens, its value is
\begin{equation}
J(\pi)=\mathbb E\Big[\mathbf 1\{a_\pi(x)\ \text{correct}\}
-\lambda\, t_\pi(x)/1000\Big],
\label{eq:objective}
\end{equation}
and the budgeted action-selection problem is $\pi^\star=\arg\max_\pi J(\pi)$ over policies respecting the caps of Definition~\ref{def:episode}.
\end{definition}

Solving Equation~\ref{eq:objective} exactly requires a model of how future vote states evolve: which answer identities further draws from $M$ would add, and what the specialist would propose. No such model is available at inference time, and estimating one on the target task would itself consume the budget being allocated. ValueFormation therefore freezes, once and on disjoint development trajectories, one-step surrogates for each action's success probability and charges every action by its expected token cost under the single price $\lambda$. The result is a sequential policy that maximizes a conservative one-step utility at every state. Neither the surrogates nor the state carries task identity, so one frozen calibration serves every task and primary model.

The distinctive action is \textsc{acquire}, and it is worth separating from verification before constructing the policy. Verifier methods can recover a correct but outvoted candidate, yet their output remains in the support of the primary pool. ValueFormation changes that support. Section~\ref{sec:mechanism} intervenes on this distinction in two ways: it removes acquisition prospectively, and it replays the same acquired traces while either keeping the primary leader or accepting every parseable proposal. The resulting ladder separates the value of obtaining new candidates from the value of integrating them correctly.

\subsection{Budgeted action selection}\label{sec:action-values}
After $k$ draws, write $\hat y_i$ for the canonicalized answer of draw $i$ when it parses. The histogram $\mathcal A$ is the vote tally
\begin{equation}
v_a(k)=\sum_{i=1}^{k}\mathbf 1\{\hat y_i=a\},\qquad
n_{\mathrm{parse}}=\sum_a v_a(k),
\label{eq:votes}
\end{equation}
so each parsed response casts exactly one vote. The leader $L_k$ is the identity with the largest count $v_1$, the runner-up has count $v_2$, $d$ counts distinct identities, and $H=-\sum_a \tfrac{v_a}{n_{\mathrm{parse}}}\log\tfrac{v_a}{n_{\mathrm{parse}}}$ is the entropy of the vote distribution --- large when votes scatter across answers and zero under unanimity. The policy observes
\begin{equation}
z_k=\Big[
\tfrac{k}{K},\;
\tfrac{n_{\mathrm{parse}}}{k},\;
\tfrac{v_1}{n_{\mathrm{parse}}},\;
\tfrac{v_2}{n_{\mathrm{parse}}},\;
\tfrac{v_1-v_2}{n_{\mathrm{parse}}},\;
\tfrac{d}{n_{\mathrm{parse}}},\;
\tfrac{H}{\log\max(n_{\mathrm{parse}},2)}
\Big].
\label{eq:state}
\end{equation}
Each component has a decision-relevant reading: $k/K$ is spent effort, how much of the sampling cap remains purchasable; $n_{\mathrm{parse}}/k$ is parse reliability, since a histogram built from few parses supports weaker vote inferences; $v_1/n_{\mathrm{parse}}$ and $(v_1-v_2)/n_{\mathrm{parse}}$ are the leader's share and margin, the primary evidence that stopping is safe; $d/n_{\mathrm{parse}}$ counts distinct identities; and the entropy ratio bounds overall disagreement in $[0,1]$, keeping early and late states comparable. The representation contains no task or dataset feature. If no draw parses, all vote-derived components are defined as zero.

For each action $a\in\{\textsc{stop},\textsc{primary},\textsc{acquire}\}$, a regularized logistic model estimates a success probability
\begin{equation}
\hat q_a(z_k)=\sigma\big(w_a^{\top} z_k+b_a\big),\qquad
\sigma(u)=\frac{1}{1+e^{-u}}.
\label{eq:logistic}
\end{equation}
The same seven features feed all three predictors, so the actions are compared on one shared representation and each contributes a single interpretable score. The targets have operational meanings. $\hat q_{\mathrm{stop}}$ predicts whether the current leader is correct; $\hat q_{\mathrm{primary}}$, fitted on nonterminal states, predicts whether plurality will be correct at the $K$-draw cap; and $\hat q_{\mathrm{acquire}}$ predicts whether a reasoning-specialist proposal is correct. With empirical mean token costs $\bar c_P$ and $\bar c_R$, the action utilities are
\begin{align}
U_{\mathrm{stop}}(z_k)&=\hat q_{\mathrm{stop}}(z_k), \nonumber\\
U_{\mathrm{primary}}(z_k)&=\hat q_{\mathrm{primary}}(z_k)
 -\lambda (K-k)\bar c_P/1000, \label{eq:utilities}\\
U_{\mathrm{acquire}}(z_k)&=\hat q_{\mathrm{acquire}}(z_k)
 -\lambda \bar c_R/1000. \nonumber
\end{align}
Writing $A_k$ for the available actions (all three, except that \textsc{primary} is removed at $k=K$), the policy selects
\begin{equation}
a^\star_k=\arg\max_{a\in A_k}\, U_a(z_k),
\label{eq:policy}
\end{equation}
breaking exact ties toward stopping. At least one primary draw is always made before any decision, so the first action choice occurs at $k=1$. A primary action draws once and recomputes all values, so Equation~\ref{eq:utilities} is a conservative receding-horizon approximation: it charges continuation as if the remaining cap were used even though a later state may stop sooner.

\paragraph{Calibration.}
The three predictors are fitted on Llama-primary trajectories from rows 32--127 of GSM8K, MATH-500, and SVAMP, disjoint from every evaluation row. These 288 development problems provide 2,880 stop states, 2,592 primary transitions, and 60 observed reasoning-proposal outcomes. All states are pooled across tasks, which is what makes the predictors task-blind by construction rather than by post hoc filtering. The measured mean costs are 466 tokens for a primary draw and 4,602 tokens for a reasoning proposal; the frozen default is $\lambda=0.0025$. Decoding and regularization settings are reported in Section~\ref{sec:evaluation}.

The predictors need only rank actions under a shared cost scale; they are not required to be calibrated probabilities on every transferred primary. The transfer experiment in Section~\ref{sec:cross-primary} keeps every coefficient and cost fixed when the primary changes to Qwen3-8B or Ministral-8B.

\paragraph{Cost monotonicity.}\label{prop:cost}
At a fixed state $z_k$ with action scores $\hat q_a-\lambda c_a$ and $c_{\mathrm{stop}}=0$, increasing the token price cannot switch the argmax to a more expensive action: if $a_1$ is optimal at $\lambda_1$ and $a_2$ is optimal at $\lambda_2>\lambda_1$, adding the two optimality inequalities gives $(\lambda_2-\lambda_1)(c_{a_1}-c_{a_2})\ge0$, hence $c_{a_2}\le c_{a_1}$. Section~\ref{sec:cost-robustness} tests the realized implication after sequential state changes: mean calls and tokens should fall as $\lambda$ rises.

\subsection{Accepting an acquired proposal}\label{sec:fusion}
Acquisition terminates primary sampling and obtains one proposal from DeepSeek-R1-Distill-Qwen-7B at temperature zero, with an 8,192-token completion cap \cite{deepseekr1}. The final answer must then be chosen between the incumbent leader and the proposal. ValueFormation makes this choice with a frozen posterior comparison that retains the incumbent unless pooled evidence favors the proposal. The comparison is an equal-weight log-opinion pool of three development source profiles.

Each profile contains a wrong-answer concentration parameter $\alpha$, proposal accuracy $\rho$, and an echo rate $\kappa$ measuring how often an incorrect proposal repeats an existing primary error rather than introduces a new identity. Let $L$ be the incumbent leader, $P$ the parsed proposal, $n=n_{\mathrm{parse}}$, and $c=\hat q_{\mathrm{stop}}(z_k)$ the incumbent-confidence prior supplied by the stop predictor; let $v_L$ and $v_P$ be the vote counts of the leader and of the pooled answer that $P$ repeats, and let $\eta=0.1$ be the frozen novel-identity mass. Profile $j=(\alpha_j,\rho_j,\kappa_j)$ fixes a proposal prior mass $m_j$ and a wrong-proposal likelihood $w_j$,
\begin{equation}
(m_j,w_j)=
\begin{cases}
\Big(\dfrac{v_P}{n-v_L},\;(1-\rho_j)\,\kappa_j\,\dfrac{v_P}{n-v_L}\Big),
& P\in\mathcal A,\\[7pt]
\Big(\dfrac{\alpha_j}{n+\alpha_j}\,\eta,\;(1-\rho_j)(1-\kappa_j)\,\eta\Big),
& P\notin\mathcal A,
\end{cases}
\label{eq:fusion-mass}
\end{equation}
and its posterior log odds in favor of the incumbent after observing the proposal are
\begin{equation}
\ell_j=\log\frac{c}{(1-c)\,m_j}-\log\frac{\rho_j}{w_j}.
\label{eq:fusion-logodds}
\end{equation}
The two cases of Equation~\ref{eq:fusion-mass} encode how much the proposal's arrival should move the incumbent's odds. A proposal repeating an identity already in the histogram ($P\in\mathcal A$) carries mass proportional to that identity's share of non-leader votes --- an answer the primary itself produced repeatedly is harder to dismiss --- while a novel identity ($P\notin\mathcal A$) is a priori implausible, its mass shrinking as the parsed pool grows. The wrong-proposal likelihood applies the same decomposition to failure: a wrong proposal most plausibly echoes one of the primary's existing errors. Equation~\ref{eq:fusion-logodds} nets the incumbent's prior odds $c/(1-c)$ against the proposal's prior mass and its wrong-to-correct likelihood ratio, so strong incumbent confidence, low proposal plausibility, or easily explained wrongness each favor retaining the leader.

The policy averages the three profiles into a pooled score
\begin{equation}
\bar\ell=\frac{1}{3}\sum_{j=1}^{3}\ell_j,
\label{eq:fusion-pool}
\end{equation}
and retains the incumbent iff $\bar\ell\ge0$, returning the proposal otherwise. If $P=L$, the leader is retained directly; a truncated or unparseable proposal carries no evidence and leaves the incumbent unchanged. The pool is task-blind: no dataset identity selects among the profiles. Supplementary Appendix~A gives the frozen profile values and the full calculation.

\subsection{Algorithm, complexity, and code route}\label{sec:algorithm}
\begin{algorithm}[t]
\caption{ValueFormation inference policy for mathematics.}
\label{alg:policy}
\begin{algorithmic}[1]
\STATE \textbf{input:} problem $x$; primary $M$; cap $K$; token price $\lambda$
\STATE $\mathcal{A}\leftarrow\emptyset$
\FOR{$k=1,\dots,K$}
  \STATE draw one response from $M$ and update answer histogram $\mathcal{A}$
  \STATE compute $z_k$ by Equation~\ref{eq:state}
  \STATE compute available utilities by Equation~\ref{eq:utilities}
  \STATE $a^\star\leftarrow\arg\max_a U_a(z_k)$
  \IF{$a^\star=\textsc{stop}$}
    \STATE \textbf{return} the primary leader
  \ELSIF{$a^\star=\textsc{acquire}$}
    \STATE obtain one reasoning proposal
    \STATE \textbf{return} posterior acceptance decision (Section~\ref{sec:fusion})
  \ENDIF
\ENDFOR
\STATE \textbf{return} the primary leader
\end{algorithmic}
\end{algorithm}

The implementation rebuilds the vote counter and entropy from the parsed prefix after each draw, giving $O(K^2)$ worst-case state computation; $K=10$ makes this negligible beside inference. Each loop iteration costs one primary call followed by constant-time scoring, so a mathematics episode issues at most $K$ primary draws and at most one specialist call. Evaluating the three linear predictors is constant-time for the fixed seven-dimensional representation, and fusion scores only the incumbent and proposal under three fixed profiles. The control computation adds no model calls. Calibration is a one-time $O(DK)$ pass over $D$ development trajectories, followed by small convex logistic fits.

\paragraph{Public-interface code route.}
Code tasks are identified from the public task interface before the mathematics loop. Literal public doctests activate execution-grounded adaptive sampling. Without such examples, the policy returns one proposal from a non-thinking Qwen3-8B code specialist. No generated hidden test, benchmark name, or gold label enters the decision, and the route is fixed across nominal primary models. The mathematics experiments carry the central acquisition claim, since only there do all three actions compete under Equation~\ref{eq:utilities}; the code route extends the same interface to a task family with executable feedback.

\section{Experiments}\label{sec:evaluation}
We organize the evaluation around four questions: whether the final unified policy improves over established alternatives, whether its Llama-fitted values transfer to other primaries, whether acquisition is load-bearing, and whether the common token price controls realized cost as predicted.

\begin{table*}[t]
\centering
\caption{Claim-population accuracy (mean $\pm$ SD, \%). Bold/underline mark best/second-best; blue marks ValueFormation. $\Delta$ best uses problem-cluster 95\% intervals.}
\label{tab:unified-transfer}
\scriptsize
\setlength{\tabcolsep}{2pt}
\begin{tabular}{lrrrrrrrrrrrr}
\toprule
 & \multicolumn{3}{c}{GSM8K} & \multicolumn{3}{c}{MATH-500} & \multicolumn{3}{c}{SVAMP} & \multicolumn{3}{c}{HumanEval$^\dagger$} \\
\cmidrule(lr){2-4}\cmidrule(lr){5-7}\cmidrule(lr){8-10}\cmidrule(lr){11-13}
Policy & Llama & Qwen & Ministral & Llama & Qwen & Ministral & Llama & Qwen & Ministral & Llama & Qwen & Ministral \\
\midrule
greedy & 86.01\,{\tiny$\pm$0.26} & 87.35\,{\tiny$\pm$0.42} & 89.22\,{\tiny$\pm$0.10} & 49.91\,{\tiny$\pm$0.41} & 56.27\,{\tiny$\pm$2.19} & 54.39\,{\tiny$\pm$1.33} & 83.91\,{\tiny$\pm$0.89} & 87.40\,{\tiny$\pm$1.68} & 88.18\,{\tiny$\pm$0.67} & 41.67\,{\tiny$\pm$0.00} & 53.70\,{\tiny$\pm$8.02} & 46.30\,{\tiny$\pm$1.60} \\
sample1 & 86.12\,{\tiny$\pm$0.65} & 87.83\,{\tiny$\pm$0.91} & 87.35\,{\tiny$\pm$0.84} & 48.57\,{\tiny$\pm$1.48} & 56.45\,{\tiny$\pm$0.93} & 53.85\,{\tiny$\pm$3.00} & 81.01\,{\tiny$\pm$1.46} & 87.40\,{\tiny$\pm$0.67} & 85.85\,{\tiny$\pm$1.46} & 42.59\,{\tiny$\pm$3.21} & 52.78\,{\tiny$\pm$2.78} & 48.15\,{\tiny$\pm$6.42} \\
SC-10 & 91.35\,{\tiny$\pm$0.75} & 91.46\,{\tiny$\pm$0.05} & 92.39\,{\tiny$\pm$0.17} & 58.51\,{\tiny$\pm$1.53} & 59.23\,{\tiny$\pm$1.09} & 61.38\,{\tiny$\pm$1.09} & 88.57\,{\tiny$\pm$2.04} & 90.50\,{\tiny$\pm$1.21} & 91.67\,{\tiny$\pm$0.67} & 47.22\,{\tiny$\pm$0.00} & 49.07\,{\tiny$\pm$1.60} & 54.63\,{\tiny$\pm$1.60} \\
ASC & 92.05\,{\tiny$\pm$0.24} & 92.84\,{\tiny$\pm$0.24} & \underline{92.78\,{\tiny$\pm$0.17}} & 61.02\,{\tiny$\pm$0.93} & 62.54\,{\tiny$\pm$1.12} & 62.90\,{\tiny$\pm$0.71} & \underline{89.15\,{\tiny$\pm$0.89}} & \underline{91.28\,{\tiny$\pm$1.01}} & \underline{91.86\,{\tiny$\pm$0.58}} & \underline{50.93\,{\tiny$\pm$4.24}} & \underline{54.63\,{\tiny$\pm$5.78}} & 51.85\,{\tiny$\pm$1.60} \\
ESC & 91.83\,{\tiny$\pm$0.26} & \underline{93.00\,{\tiny$\pm$0.13}} & 92.50\,{\tiny$\pm$0.10} & \underline{62.46\,{\tiny$\pm$0.56}} & 61.65\,{\tiny$\pm$1.33} & \underline{64.25\,{\tiny$\pm$0.71}} & \underline{89.15\,{\tiny$\pm$0.67}} & \underline{91.28\,{\tiny$\pm$0.58}} & \textbf{92.05\,{\tiny$\pm$0.34}} & 48.15\,{\tiny$\pm$3.21} & 51.85\,{\tiny$\pm$4.24} & 51.85\,{\tiny$\pm$3.21} \\
USC & 89.25\,{\tiny$\pm$0.17} & 90.88\,{\tiny$\pm$0.55} & 89.73\,{\tiny$\pm$0.27} & 53.94\,{\tiny$\pm$4.50} & \underline{63.35\,{\tiny$\pm$0.41}} & 58.69\,{\tiny$\pm$3.02} & 84.50\,{\tiny$\pm$1.21} & 89.15\,{\tiny$\pm$0.89} & 89.73\,{\tiny$\pm$0.89} & 46.30\,{\tiny$\pm$5.78} & 48.15\,{\tiny$\pm$4.24} & \underline{58.33\,{\tiny$\pm$4.81}} \\
GDV & 92.05\,{\tiny$\pm$0.35} & 91.46\,{\tiny$\pm$0.35} & 91.86\,{\tiny$\pm$0.25} & 59.86\,{\tiny$\pm$2.29} & 59.86\,{\tiny$\pm$0.94} & 60.30\,{\tiny$\pm$0.86} & 88.18\,{\tiny$\pm$0.89} & 90.12\,{\tiny$\pm$1.54} & 90.89\,{\tiny$\pm$1.21} & 42.59\,{\tiny$\pm$1.60} & 51.85\,{\tiny$\pm$5.78} & 50.93\,{\tiny$\pm$4.24} \\
ReASC & \underline{92.28\,{\tiny$\pm$0.08}} & 91.83\,{\tiny$\pm$0.48} & 92.05\,{\tiny$\pm$0.51} & 60.39\,{\tiny$\pm$0.78} & 60.04\,{\tiny$\pm$0.41} & 60.13\,{\tiny$\pm$2.24} & \textbf{90.12\,{\tiny$\pm$0.58}} & 90.12\,{\tiny$\pm$1.01} & 91.47\,{\tiny$\pm$0.89} & 50.00\,{\tiny$\pm$5.56} & 49.07\,{\tiny$\pm$5.78} & 52.78\,{\tiny$\pm$2.78} \\
\ourrowcolor\emph{ValueFormation} & \textbf{92.67\,{\tiny$\pm$0.21}} & \textbf{94.85\,{\tiny$\pm$0.34}} & \textbf{93.09\,{\tiny$\pm$0.56}} & \textbf{80.29\,{\tiny$\pm$1.53}} & \textbf{84.14\,{\tiny$\pm$0.97}} & \textbf{79.84\,{\tiny$\pm$1.76}} & \underline{89.15\,{\tiny$\pm$0.67}} & \textbf{91.67\,{\tiny$\pm$0.67}} & 91.28\,{\tiny$\pm$0.58} & \textbf{71.30\,{\tiny$\pm$1.60}} & \textbf{71.30\,{\tiny$\pm$1.60}} & \textbf{71.30\,{\tiny$\pm$1.60}} \\
$\Delta$ best (95\% CI) & \begin{tabular}[c]{@{}c@{}}+0.39\\[-2pt]{\tiny[-0.42, +1.20]}\end{tabular} & \begin{tabular}[c]{@{}c@{}}+1.85\\[-2pt]{\tiny[+0.90, +2.85]}\end{tabular} & \begin{tabular}[c]{@{}c@{}}+0.31\\[-2pt]{\tiny[-0.50, +1.12]}\end{tabular} & \begin{tabular}[c]{@{}c@{}}+17.83\\[-2pt]{\tiny[+13.17, +22.49]}\end{tabular} & \begin{tabular}[c]{@{}c@{}}+20.79\\[-2pt]{\tiny[+16.31, +25.27]}\end{tabular} & \begin{tabular}[c]{@{}c@{}}+15.59\\[-2pt]{\tiny[+11.29, +19.89]}\end{tabular} & \begin{tabular}[c]{@{}c@{}}-0.97\\[-2pt]{\tiny[-3.88, +1.94]}\end{tabular} & \begin{tabular}[c]{@{}c@{}}+0.39\\[-2pt]{\tiny[-2.52, +3.29]}\end{tabular} & \begin{tabular}[c]{@{}c@{}}-0.78\\[-2pt]{\tiny[-2.52, +0.78]}\end{tabular} & \begin{tabular}[c]{@{}c@{}}+20.37\\[-2pt]{\tiny[+5.56, +36.11]}\end{tabular} & \begin{tabular}[c]{@{}c@{}}+16.67\\[-2pt]{\tiny[-0.93, +34.26]}\end{tabular} & \begin{tabular}[c]{@{}c@{}}+12.96\\[-2pt]{\tiny[-3.70, +29.63]}\end{tabular} \\
\bottomrule
\end{tabular}

\end{table*}

\subsection{Experimental setup}\label{sec:setup}
\paragraph{Tasks and population.}
We use GSM8K \cite{cobbe2021math}, MATH-500 \cite{hendrycks2021math,lightman2024verify}, SVAMP \cite{patel2021svamp}, and HumanEval \cite{chen2021humaneval}. Calibration uses mathematics rows 32--127. Evaluation uses every public-test row from index 128 onward: 1,191/372/172/36 problems. The three fixed seeds are shared across the final-policy conditions. The code-breadth analysis additionally evaluates the 257-problem sanitized MBPP suite \cite{austin2021program} with official tests.

\paragraph{Models and budgets.}
The primary is Llama-3.1-8B-In\-struct, Qwen3-8B, or Ministral-8B-Instruct. Qwen runs in non-thinking mode. Mathematics acquisition uses DeepSeek-R1-Distill-Qwen-7B; the code route uses non-thinking Qwen3-8B. Three further models appear only in the transfer, stress, breadth, and payload-swap analyses: the larger primary Qwen3-14B (non-thinking; run in thinking mode it also supplies an equal-strength acquisition payload), the weaker payload Qwen3-4B-Thinking-2507, and the weak stress specialist DeepSeek-R1-Distill-Qwen-1.5B. The hard per-problem limits are 13 calls, 40,960 completion tokens, and 98,304 prompt tokens. Development and evaluation trajectories draw from the primary at temperature 0.7 with top-$p$ 0.95 and a 2,048-token completion cap; the logistic predictors use standardized features, an unpenalized intercept, and $\ell_2$ regularization with strengths 2, 2, and 4 for the stop, primary, and acquisition models. Unless varied explicitly, $\lambda=0.0025$.

\paragraph{Baselines and uncertainty.}
The frozen comparison bundle contains greedy, one-sample, SC-10, ASC, ESC, USC, GDV, and ReASC under three seeds. For each primary and task, we report the highest baseline accuracy after restricting its saved per-problem outcomes to the exact evaluation IDs. Intervals are 20,000-replicate problem-cluster bootstraps of the difference between the two three-seed mean correctness vectors, retaining all seed outcomes per sampled problem, so they quantify problem-population uncertainty rather than treating seeds as independent examples. Baseline usage is recomputed per evaluation ID from the frozen per-problem traces (median over the three seeds), so reported baseline costs are exact-subset costs on the same rows as the policy.

\subsection{RQ1: Does the unified policy improve the claim population?}\label{sec:main-results}
Table~\ref{tab:unified-transfer} reports the claim-population comparison. On MATH-500, ValueFormation reaches 80.29\%, 17.83 points above ESC, with a 95\% interval of $[+13.17,+22.49]$. HumanEval improves by 20.37 points over ASC on the fixed code route; with 36 problems this serves as breadth evidence, and the mechanism claims below rest on the mathematics tasks. GSM8K improves by 0.39 points, with an interval that includes zero, while SVAMP is neutral: the Llama/Qwen/Ministral deltas are $-0.97$, $+0.39$, and $-0.78$, and all three intervals include zero.

The margin on MATH-500 lies outside the range that the choice of baseline can explain. Across the eight retained policies, MATH-500 accuracy spans 13.89 points, from 48.57 (one-sample) to 62.46 (ESC); the 17.83-point margin over the strongest of the eight exceeds that entire spread. On GSM8K the arithmetic reverses: the baselines span only 6.27 points and cluster near 92, and the +0.39 gain sits inside that band. This matches the coverage account of Section~\ref{sec:problem}: where ten primary draws usually contain the correct answer, selection alone is nearly sufficient, whereas on MATH-500 the incumbent candidate set is often missing the answer that acquisition can supply --- an interpretation the mechanism experiment below tests directly.

This pattern is useful because it separates a performance claim from a mechanism claim. A policy that merely improves selection should help most where the correct answer is already present but hard to identify. Instead, the largest gains appear on the task where the retained baselines differ widely yet still plateau far below the specialist-augmented policy. The small GSM8K and SVAMP deltas are therefore the negative side of the same hypothesis: when primary coverage and existing selection already approach saturation, there is little missing support for acquisition to add.

The Llama policy uses 4.60/4.69/4.27 calls on GSM8K/\allowbreak MATH-500/SVAMP (token means 2,021/5,226/2,159; HumanEval 1.03 calls, 282 tokens) despite allowing ten primary draws and one acquisition: the rule preserves a large benefit where proposal coverage is scarce.

\looseness=-1
The code-route advantage also holds in breadth: on the 257-problem MBPP suite under free routing (supplementary Appendix~F), ValueFormation attains the highest accuracy of nine methods, 74.19\%, at one call and about 40 tokens, with intervals excluding zero against six of eight baselines ($+5.71$ to $+6.36$) and the two unresolved methods at nine and eleven calls.

\subsection{RQ2: Does one calibration transfer across primaries?}\label{sec:cross-primary}
Across primaries in Table~\ref{tab:unified-transfer}, the strongest transfer pattern is on MATH-500: the same frozen policy improves over the best baseline by 17.83 points for Llama, 20.79 for Qwen, and 15.59 for Ministral; all three intervals exclude zero. Qwen also gains 1.85 points on GSM8K with a positive interval. Llama and Ministral show small, unresolved GSM8K gains, and all three SVAMP deltas are unresolved. The MATH margins are not an artifact of weak baseline ceilings: Ministral's strongest baseline (ESC, 64.25) is the highest of the three primaries, yet its gain is still 15.59 points, and the smallest interval lower bound across primaries ($+11.29$) exceeds every non-MATH gain in the table. In absolute terms Qwen benefits most, reaching 84.14 on MATH-500 against its own best baseline of 63.35.

Model identity changes while the full policy does not: the frozen package retains a recurring MATH advantage across primaries, with smaller and less uniformly resolved gains on high-coverage mathematics tasks. Across the nine mathematics settings the transferred policy uses 4.18--4.86 calls and 1,692--5,226 tokens per problem (supplementary Appendix~C), retaining the cost profile of the Llama campaign. On HumanEval the transfer is exact by construction: the fixed code route resolves 35 of 36 problems through the shared code specialist without primary sampling, so both transferred primaries reproduce the Llama per-seed accuracies (mean 71.30 at 1.03 calls), $+16.67$ $[-0.93,+34.26]$ over the best Qwen baseline and $+12.96$ $[-3.70,+29.63]$ over the best Ministral baseline; with 36 problems these intervals include zero, so the HumanEval transfer remains breadth evidence.

The cross-primary result also constrains what can explain the MATH gains. The policy receives no primary-model identifier, and the value models are not refit when answer style, token length, and baseline strength change. If the learned rule were only exploiting idiosyncrasies of Llama development traces, the transferred rows would be expected to lose both accuracy and cost discipline. Instead, the same qualitative operating point recurs: a few primary samples are collected, acquisition is purchased on difficult states, and the final cost stays far below exhaustive self-consistency or verifier-heavy baselines. This is the main evidence that the state representation captures a portable signal rather than a memorized task route.

\looseness=-1
The same untouched policy transfers to a fourth, larger primary: on Qwen3-14B (non-thinking) over the identical claim rows it reaches 95.83 GSM8K and 84.86 MATH-500 (three seeds) at 4.20/4.26 calls and 1,849/4,479 tokens, beating the unconditional R1-only control by $+7.33$ $[+5.74,+8.98]$ and $+3.67$ $[+1.08,+6.36]$ (supplementary Appendix~F). With this stronger primary the conditional policy exceeds the always-on specialist on both tasks, and the 84.86 exceeds every Qwen3-8B exact-ID baseline by more than 21 points.

\subsection{RQ3: Is acquisition load-bearing?}\label{sec:mechanism}
Removing acquisition reduces Llama MATH accuracy from 80.29\% to 59.86\%, a 20.43-point effect with interval $[+16.85,+24.19]$, while GSM8K falls by 1.40 points $[+0.62,+2.21]$. The complementary R1-only control clarifies the mechanism: one greedy specialist call scores 81.18 on MATH-500, statistically tied with the Llama policy, but only 88.50 on GSM8K, 4.17 points below the policy. Acquisition is therefore load-bearing when the primary candidate set lacks coverage; when primary coverage is already high, the value is conditional allocation rather than always-on specialization.

\looseness=-1
A coverage cross-tabulation locates where the rule spends. On MATH-500 the independent no-acquisition pool lacks a correct candidate on 28.85\% of rows; the policy acquires on 85.09\% of those rows, versus 40.81\% when the pool already contains the answer, and fires on 94.67\% of rows where the specialist supplies the missing answer. GSM8K shows the same direction at lower rate because 96.31\% of pools already contain the answer (supplementary Appendix~G).

\begin{figure}[t]
\centering
\includegraphics[width=\columnwidth]{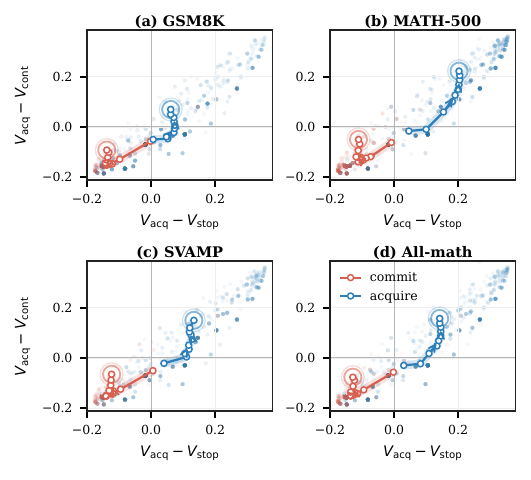}
\caption{Inference trajectories in value-state space. Points are primary-sampling prefixes projected onto $V_{\mathrm{acq}}-V_{\mathrm{stop}}$ and $V_{\mathrm{acq}}-V_{\mathrm{cont}}$; arrows show draw-index direction. Blue trajectories acquire; red trajectories commit.}
\Description{Four value-state trajectory plots for GSM8K, MATH-500, SVAMP, and pooled mathematics. Blue trajectories move into the region where acquisition has positive value margin over stopping and continuing, while red commit trajectories remain closer to the non-acquisition region.}
\label{fig:value-state-trajectories}
\end{figure}

\begin{figure}[t]
\centering
\includegraphics[width=\columnwidth]{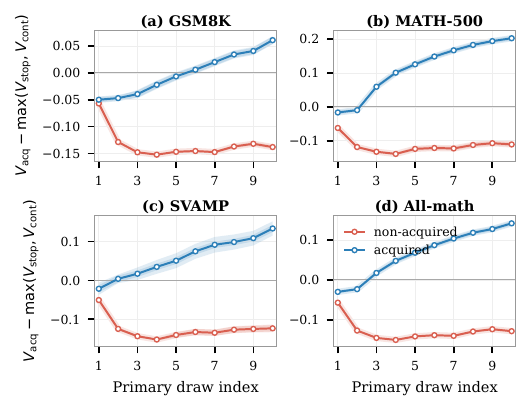}
\caption{Inference dynamics over primary-sampling prefixes. Curves show mean acquisition advantage, $V_{\mathrm{acq}}-\max(V_{\mathrm{stop}},V_{\mathrm{cont}})$, with $\pm$ one standard error. Blue trajectories acquire; red trajectories commit.}
\Description{Four point-and-line plots for GSM8K, MATH-500, SVAMP, and pooled mathematics. The blue acquired trajectories rise above the red non-acquired trajectories in acquisition advantage as primary draw index increases.}
\label{fig:inference-dynamics}
\end{figure}

Figures~\ref{fig:value-state-trajectories} and~\ref{fig:inference-dynamics} visualize the same allocation geometry. Trajectories that eventually acquire move into the region where acquisition beats both stopping and continuing, and their draw-index curves cross into positive acquisition advantage as evidence accumulates. These plots are diagnostics; the effect sizes come from the ablations and replays.

The diagnostics are useful for ruling out two simpler readings. The policy does not acquire immediately on every difficult-looking problem; it waits for the primary pool to reveal whether the leader is stable, whether alternatives are accumulating, and whether another same-model draw is likely to resolve the state. Acquired trajectories therefore mark a movement from ambiguity toward positive acquisition margin, not a static route chosen from the prompt. Conversely, commit trajectories include both confident stops and states where continuing the primary remains more valuable than buying the specialist. This is the behavior expected from a three-action value policy rather than from a binary specialist detector.

\begin{figure}[t]
\centering
\includegraphics[width=\columnwidth]{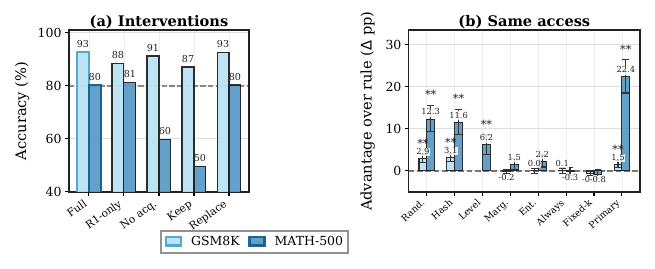}
\caption{Acquisition interventions and same-access rules. Left: Llama-primary accuracies. Right: ValueFormation advantage over each shared-payload rule. Intervals are problem-cluster 95\% CIs; $^{*}$/$^{**}$ denote max-$T<0.05/0.01$.}
\Description{Two-panel vertical grouped bar figure. The left panel compares mechanism intervention accuracies for GSM8K and MATH-500, showing large MATH-500 loss without acquisition and recovery under replace-on-parse. The right panel compares ValueFormation against same-access acquisition rules, showing large positive MATH-500 advantages over random, hash, level routing, and primary-only rules.}
\label{fig:rq3-mechanism-bars}
\end{figure}

\looseness=-1
Figure~\ref{fig:rq3-mechanism-bars} turns this into a compact intervention summary. Panel (a) shows that removing acquisition creates the large MATH gap and that replace-on-parse recovers the full policy on acquired traces; panel (b) gives every alternative rule the same trajectories, continuation access, specialist payload, and posterior integrator. The shared-payload policy anchor is 93.06/81.81 on GSM8K/MATH-500; rule scalars are calibrated on a disjoint hash split and the 15-comparison family is corrected by Holm and per-dataset Westfall--Young max-$T$ (supplementary Appendix~G).

The state-conditioned rule is strongest where coverage is scarce. On MATH-500 it beats matched-rate random acquisition by 12.31 points $[+9.32,+15.50]$, identity-hash selection by 11.56, level routing by 6.18, and primary-only by 22.40; these survive the corrected max-$T$ tests. On GSM8K the corrected wins are smaller, against random (+2.92), hash (+3.13), and primary-only (+1.51), while margin/entropy thresholds tie it and always-acquire matches accuracy at 35\% more tokens. The rule-level account is thus causal: scarce-coverage rows require state-conditioned selection, whereas saturated rows mainly require cheap allocation.

Two alternatives are especially important. Always acquiring tests whether the specialist should simply be called by default; it is competitive only when the primary is already strong enough that extra calls rarely change the answer, and it spends more tokens in that regime. Primary-only tests the opposite story, that enough same-model samples substitute for heterogeneity; it remains far behind on MATH-500 under the same trace budget. The useful behavior is therefore neither ``always collaborate'' nor ``sample deeper'', but a switch conditioned on the current pool's evidence.

\looseness=-1
Fixed-trajectory replays isolate the acquired answer from the choice to acquire. On MATH-500, keeping the primary leader on acquired traces scores 49.55\%, whereas replacing with every complete parseable proposal reaches 80.29\%, exactly matching the full policy; GSM8K shows the same equality. These acceptance decisions are not fragile: 1,114 genuine proposal comparisons have median pooled log-odds margin $-3.20$ and at most one decision flips under 20\% profile perturbations. Under a much weaker specialist, the posterior rule still beats keep-leader in all six cells, while an equal-strength Qwen3-14B payload replays MATH-500 to 81.36 versus the deployed 81.81. Finally, refitting the acquisition predictor on 1,790 rebuilt outcomes changes only 6.85\% of visited actions, retains all realized acquisitions, improves pooled MATH-500 by $+2.05$ $[+0.49,+3.71]$, and reduces mean calls. The acquisition predictor scores acquisition against stopping and continuation, not only against specialist correctness; a probe-only threshold rule that does the latter improves selected rows but withholds from weak-primary rows and trails the policy overall.

\subsection{RQ4: Does the token price control realized cost?}\label{sec:cost-robustness}
\begin{table}[t]
\centering
\caption{Cost-weight sensitivity with a Llama primary. Accuracy is mean $\pm$ SD; tokens are per problem.}
\label{tab:unified-cost}
\small
\begin{tabular}{lrrrr}
\toprule
Task & $\lambda$ & Accuracy & Calls & Tokens / R1 share \\
\midrule
GSM8K & 0 & 92.84 $\pm$ 0.10 & 6.14 & 2,586 / 15.59\% \\
 & 0.0025 & 92.67 $\pm$ 0.21 & 4.60 & 2,021 / 15.00\% \\
 & 0.005 & 92.50 $\pm$ 0.46 & 2.70 & 1,314 / 12.90\% \\
\midrule
MATH-500 & 0 & 80.02 $\pm$ 0.68 & 5.38 & 5,656 / 55.82\% \\
 & 0.0025 & 80.29 $\pm$ 1.53 & 4.69 & 5,226 / 53.58\% \\
 & 0.005 & 76.79 $\pm$ 1.48 & 3.43 & 4,375 / 49.19\% \\
\bottomrule
\end{tabular}

\end{table}

\begin{table*}[t]
\centering
\caption{Inference cost per problem (calls/tokens). Blue marks ValueFormation; baselines use full-test means.}
\label{tab:cost-context}
\scriptsize
\setlength{\tabcolsep}{3pt}
\begin{tabular}{lcccccccccccc}
\toprule
 & \multicolumn{3}{c}{GSM8K} & \multicolumn{3}{c}{MATH-500} & \multicolumn{3}{c}{SVAMP} & \multicolumn{3}{c}{HumanEval$^\dagger$} \\
\cmidrule(lr){2-4}\cmidrule(lr){5-7}\cmidrule(lr){8-10}\cmidrule(lr){11-13}
Policy & Llama & Qwen & Ministral & Llama & Qwen & Ministral & Llama & Qwen & Ministral & Llama & Qwen & Ministral \\
\midrule
greedy & 1.00/364 & 1.00/1,740 & 1.00/399 & 1.00/767 & 1.00/2,094 & 1.00/641 & 1.00/331 & 1.00/1,531 & 1.00/323 & 1.00/283 & 1.00/1,755 & 1.00/252 \\
sample1 & 1.00/365 & 1.00/1,757 & 1.00/398 & 1.00/774 & 1.00/2,099 & 1.00/617 & 1.00/320 & 1.00/1,545 & 1.00/309 & 1.00/288 & 1.00/1,766 & 1.00/247 \\
SC-10 & 10.00/3,658 & 10.00/17,557 & 10.00/3,984 & 10.00/7,771 & 10.00/20,975 & 10.00/6,132 & 10.00/3,154 & 10.00/15,458 & 10.00/3,092 & 11.00/3,502 & 11.00/18,644 & 11.00/3,112 \\
ASC & 8.03/3,401 & 10.51/21,019 & 7.35/3,170 & 20.15/19,644 & 24.65/53,650 & 18.99/14,095 & 8.33/3,581 & 10.70/20,547 & 6.36/2,264 & 10.29/3,514 & 5.27/8,659 & 10.56/3,144 \\
ESC & 8.07/3,450 & 14.14/28,672 & 6.98/3,038 & 21.03/20,838 & 28.93/62,812 & 19.86/14,849 & 8.82/3,981 & 12.65/24,638 & 6.44/2,326 & 10.37/3,598 & 5.28/8,640 & 10.68/3,155 \\
USC & 9.00/5,038 & 9.00/27,673 & 9.00/5,634 & 9.00/11,476 & 9.00/33,354 & 9.00/9,008 & 9.00/4,339 & 9.00/24,430 & 9.00/4,326 & 9.00/3,288 & 9.00/27,468 & 9.00/2,809 \\
GDV & 10.06/3,759 & 10.05/17,658 & 10.05/4,024 & 10.34/8,677 & 10.07/21,123 & 10.31/6,556 & 10.08/3,284 & 10.03/15,557 & 10.03/3,112 & 11.00/5,087 & 11.00/34,002 & 11.00/3,325 \\
ReASC & 5.18/4,233 & 6.24/23,787 & 4.25/3,637 & 8.98/17,152 & 11.19/48,265 & 8.32/12,199 & 4.65/4,053 & 6.01/21,230 & 3.67/2,564 & 6.61/4,093 & 4.29/13,686 & 6.14/3,325 \\
\ourrowcolor\emph{ValueFormation} & 4.60/2,021 & 4.29/1,981 & 4.54/2,066 & 4.69/5,226 & 4.18/4,584 & 4.86/4,830 & 4.27/2,159 & 4.25/1,692 & 4.44/1,717 & 1.03/282 & 1.03/282 & 1.03/281 \\
\bottomrule
\end{tabular}

\end{table*}

The realized costs follow the cost-monotonicity argument in Section~\ref{sec:action-values}. Raising $\lambda$ from zero to 0.005 reduces mean calls from 6.14 to 2.70 on GSM8K and from 5.38 to 3.43 on MATH-500, and token use falls at every step on both tasks (2,586/2,021/1,314 and 5,656/5,226/4,375). Accuracy responds unevenly. GSM8K varies by only 0.34 points across the sweep. MATH-500 is slightly more accurate at the middle price than at zero (80.29 versus 80.02) while saving 13\% of calls and 8\% of tokens, and the high price trades a further 3.50 points for another 16\% of MATH tokens.

The acquisition share changes less than the number of primary draws: across the sweep the R1 share falls by 2.69 points on GSM8K and 6.63 on MATH-500, while mean calls fall by 56\% and 36\%. The price therefore acts mainly on continuation, with a modest effect on whether to buy R1, consistent with acquisition being where the policy's marginal value concentrates. One utility scale thus traces a family of operating points without task-specific routing rules.

\looseness=-1
On the identical claim rows the strongest Llama GSM8K baseline (ReASC) reaches 92.28 at 2,105 tokens and 5.17 calls, where the policy scores 92.67 at 2,021; on Llama MATH-500 the strongest baseline (ESC) needs 21,319 tokens and 21.6 calls for 62.46, against the policy's 80.29 at 5,226 --- 4.1 times cheaper. Table~\ref{tab:cost-context} situates these exact-subset policy costs against every retained policy's full-test usage for all three primaries. The MATH premium is compositional: the 5,226 tokens split as 2,919 primary plus 2,307 specialist at the 53.58\% acquisition rate.

The cost table also shows why calls alone are an incomplete budget measure. Verification-heavy baselines may use few generation calls but spend many accounted tokens through long prompts, repeated completions, or teacher-forcing passes. ValueFormation's cost advantage comes from shortening both parts of the computation: it stops before the ten-draw cap on easy states and reserves expensive specialist calls for states where the predicted value of new support is high enough to pay for them. The relevant operating point is therefore not the cheapest possible run, but the frontier point that retains the MATH gain while avoiding the cost of applying the specialist or a verifier to every problem.

This matters for deployment because the same accuracy can imply different usage profiles. Always acquiring fixes the collaboration pattern before observing problem-specific evidence; primary-only deepening keeps the system simple but pays for correlated samples on cases already decided by the early vote state. The value formulation exposes a single control knob: increasing $\lambda$ moves the operating point toward shorter primary prefixes and fewer specialist calls while leaving the learned predictors unchanged.

Taken together, the four research questions give a consistent failure analysis rather than four independent wins. RQ1 establishes where the policy changes accuracy, RQ2 shows that the useful signal is not tied to one primary model, RQ3 identifies acquisition as the load-bearing action in scarce-coverage rows, and RQ4 shows that the same utility can be priced to control cost. The weak and equal-strength payload replays further delimit the claim: ValueFormation is not a promise that any second model helps, nor that acquisition should be used uniformly. It helps when the external proposal is sufficiently different and when the observed primary pool makes new support more valuable than another correlated draw. That is the operational distinction between adaptive team formation and fixed model routing.

The remaining negative cases are part of the same picture. GSM8K and SVAMP are not strong acquisition benchmarks in this setup because the strongest primary-side baselines already sit near the policy, so stronger selection and earlier stopping dominate the budget calculus. HumanEval and MBPP are reported as interface breadth rather than adaptive-acquisition evidence because the public code route fixes the specialist decision before the mathematics loop begins. These boundaries keep the empirical claim narrow: the adaptive action policy is supported where mathematics coverage is scarce, while code tasks show that the same system interface can accommodate a different evidence source.

This reading also explains why the cost result should not be reduced to a raw efficiency claim. The policy avoids paying for redundant same-model evidence while preserving the option to buy a qualitatively different proposal when the current pool has become uninformative. On saturated tasks, it behaves like a conservative stopping rule with occasional acquisition; on MATH-500, the same rule spends more specialist tokens because missing support is much more likely. A fixed router cannot express this distinction after observing the sampled candidates, and a pure sampler cannot change the source of support. The experiment section therefore supports a specific allocation claim rather than a general claim that collaboration is always cheaper or always more accurate.

\paragraph{Validity and boundaries.}
All 57 registered conditions completed with coverage one and zero model-call errors, totaling 31,338 evaluated problem instances; the specialist controls, Qwen3-14B transfer, randomized-state probe, weak-specialist replay, MBPP breadth run, complementary-payload replays, and HumanEval transfer also completed without call errors. The claim population is bounded by construction: calibration and evaluation use disjoint rows of related benchmark families, and the code route is fixed capability routing rather than adaptive code acquisition. The acquisition predictor is selection-conditioned, with randomized-state support at AUC 0.653 and 0.636 under a held-out refit. Two reporting asymmetries are worth noting: MATH-500 accuracy carries a token premium over one unconditional specialist call, and Figure~\ref{fig:rq3-mechanism-bars}b's rule family is multiplicity-corrected while Table~\ref{tab:unified-transfer}'s best-baseline selections are not.

\section{Conclusion}\label{sec:conclusion}
This paper treated test-time collaboration as a value-conditioned choice among stopping, continuing one model, and acquiring a heterogeneous proposal. The starting observation was that selection and coverage are different bottlenecks: voting and verification reweight what the candidate pool contains, and neither can recover an answer that never enters it. We found that the vote state itself signals which regime a problem is in, and that one frozen, task-blind calibration is enough to act on that signal.

The resulting policy transfers a large MATH-500 advantage across Llama, Qwen, and Ministral primaries and, unchanged, to Qwen3-14B; the fixed code route also reaches the top MBPP accuracy at the lowest interaction cost. Three interventions tie the gain to the mechanism rather than to specialist access: removing acquisition costs 20.4 Llama MATH points, fixed-trace replay attributes the gain to accepted external proposals, and a same-access replay with identical payload, integrator, and continuation access still favors the state-conditioned rule by 12.3 points over matched-rate random acquisition. A single token price traces cheaper operating points without retraining.

Overall, the results suggest pricing each test-time call by what it can add to the current candidate set rather than committing collaborators in advance. The boundary is equally clear: when primary coverage saturates, acquisition stops paying and value reduces to cheaper allocation.


\clearpage
\bibliographystyle{ACM-Reference-Format}
\bibliography{references}

\end{document}